\documentclass{article}

 \usepackage[preprint]{neurips_2026}
 \usepackage{amsmath}
  \usepackage{graphicx}
  \usepackage{subcaption}
  \usepackage[most]{tcolorbox}
\usepackage{fancyvrb}
\usepackage{fvextra}

\usepackage[utf8]{inputenc} 
\usepackage[T1]{fontenc}    
\usepackage{hyperref}       
\usepackage{url}            
\usepackage{booktabs}       
\usepackage{amsfonts}       
\usepackage{nicefrac}       
\usepackage{microtype}      
\usepackage{xcolor}         

\title{Towards a Reliable and Practical  Eval Pipeline}

  \author{%
  Emma Thuong Nguyen \\
    Salesforce \\
    \texttt{emmanguyen@salesforce.com} \\
    \And
    Abhishek Ghose \\
    Salesforce \\
    \texttt{aghose@salesforce.com} \\  
  }

\begin{document}

\maketitle

\begin{abstract}
LLM-based software systems increasingly require effective “evals” as quality gates in the development lifecycle. However, existing work typically addresses individual aspects of eval reliability rather than the full set of practical requirements. We present an end-to-end eval pipeline that combines eval checklist creation, with learned aggregation for checklist responses, to improve agreement across LLM judges and accuracy against human judgments. The framework additionally provides self-consistency, explanations, and prediction uncertainty, and we empirically demonstrate its effectiveness.
\end{abstract}

\section{Introduction}
Over the past few years Large Language Models (LLM) have become a common ingredient in software, where they assist users in a variety of tasks such as drafting emails, writing and summarizing documents. This has engendered novel challenges in the software \emph{development} process, where such product features are required to be released only after passing through rigorous quality tests. Here, the traditional practice of writing purely programmatic tests no longer suffices, and this step needs to utilize LLMs as well. For example, if a particular LLM-driven feature produces summaries, a corresponding test that checks for its \emph{coherence} or \emph{fluency} also typically needs to use an LLM.

This creates an interesting problem: the fundamental behaviors we want our tests to check, e.g., effects of LLM non-determinism, are also possessed by the tests themselves. How do we then ensure that our tests precisely measure what we intend them to, and also provide guarantees similar to a traditional quality assurance framework? 

Effective \emph{LLM-as-judge} evaluations (``evals'' henceforth) are a topic of active research, e.g., \emph{G-Eval} \citep{geval}, \emph{CheckEval} \citep{checkeval}, \emph{Prometheus-2} \citep{prometheus2}, \emph{FActScore} \citep{min-etal-2023-factscore}, dynamic rubrics based on \emph{information gain} \citep{xu2026beyond}, hybrid judges for image caption evaluation \citep{matsuda-etal-2025-vela}, \emph{GraphJudge} for evaluating knowledge graphs \citep{graphjudge}. However, most research focuses on specific aspects of evals; usually certain forms of alignment with human judges. We believe that this state of the current literature leaves open many questions wrt practical deployment .

These gaps exist both in form of  (a) not examining different aspects of alignment, and (b) not covering the full gamut of practical use-cases. In a sense, recent research leans towards addressing challenges that may be deemed \emph{necessary}, but are not \emph{sufficient} for a practical setup. Our focus is the latter.

Our primary contributions are: (a) first, we present the desiderata of a practical eval framework, outlining risks and benefits, and (b) then, we propose such a framework, and empirically show that it effectively satisfies those needs. Rather than a single novel idea, this work offers an assembly of strategies to  create an end-to-end eval pipeline that can augment a standard quality assurance pipeline.

The paper is organized as follows: we begin by listing the desiderata for a practical eval framework in \S \ref{sec:practical_eval_pipeline}, and then we present an overview of our framework in \S \ref{sec:overview}. Sections \ref{sec:experiments} -  \ref{sec:conformal_predictions} present our empirical analyses. Our final section, \S \ref{sec:conclusions}, discusses   limitations of the current study, and our conclusions.

\section{A Practical Eval Pipeline}
\label{sec:practical_eval_pipeline}

We believe that a practical eval system must at least ensure the following properties: 
\begin{itemize}

    \item[\textbf{P1.}] \textbf{Inter-LLM Agreement} (or \emph{agreement} in short): An eval output must exhibit minimal variance across LLMs. The lack of this property is a \emph{business risk}. \texttt{LLM A} might clear all tests today, but when evals are switched to use  \texttt{LLM B} (maybe due to an organization's policy) that's a harsher critic, we might end up with high failure rates.
    \item[\textbf{P2.}] \textbf{Accuracy}: Eval outcomes must closely match human judgments. This is in contrast to most empirical analyses where some form of correlation, e.g., \emph{Kendall's} $\tau$ or \emph{Spearman correlation coefficient} $\rho$, is measured, e.g., \citet{checkeval,geval}. Product quality gates are almost always hard thresholds, e.g., ``the coherence score for this summary must be $\geq 4$''. 
     For such scenarios, there is a \emph{budgeting} risk in productizing research with reported success in correlation-type metrics.

    \item[\textbf{P3.}] \textbf{Self-consistency}: An eval output should exhibit minimal variation across multiple executions against the same LLM. This directly impacts how many times a test needs to be run for it to certify that a feature works. Thus, this impacts \emph{development cost} and \emph{time-to-ship}.

    \item[\textbf{P4.}] \textbf{Explainability}: In traditional software testing, it is easy to trace  failures to pinpoint a cause, by looking at a \emph{stack trace} or \emph{logs}. This is difficult when evals are complex and are written as large text prompts. What's the equivalent of a stack trace in this new world?

    \item[\textbf{P5.}] \textbf{Confidence scores}: Since LLM-based evals are probabilistic, we need to provide some representation of \emph{confidence} for eval outcomes. We need to move away from hard predictions, e.g., \emph{``the eval score is 5''} to soft estimates, e.g., \emph{``the eval score has a 90\% prediction interval of 3-5''}. 

\end{itemize}
\textbf{P1}-\textbf{P3} characterize forms of alignment, while \textbf{P4} and \textbf{P5} represent additional critical use-cases.

  \begin{figure}[t!]
      \centering
      \includegraphics[width=1\linewidth]{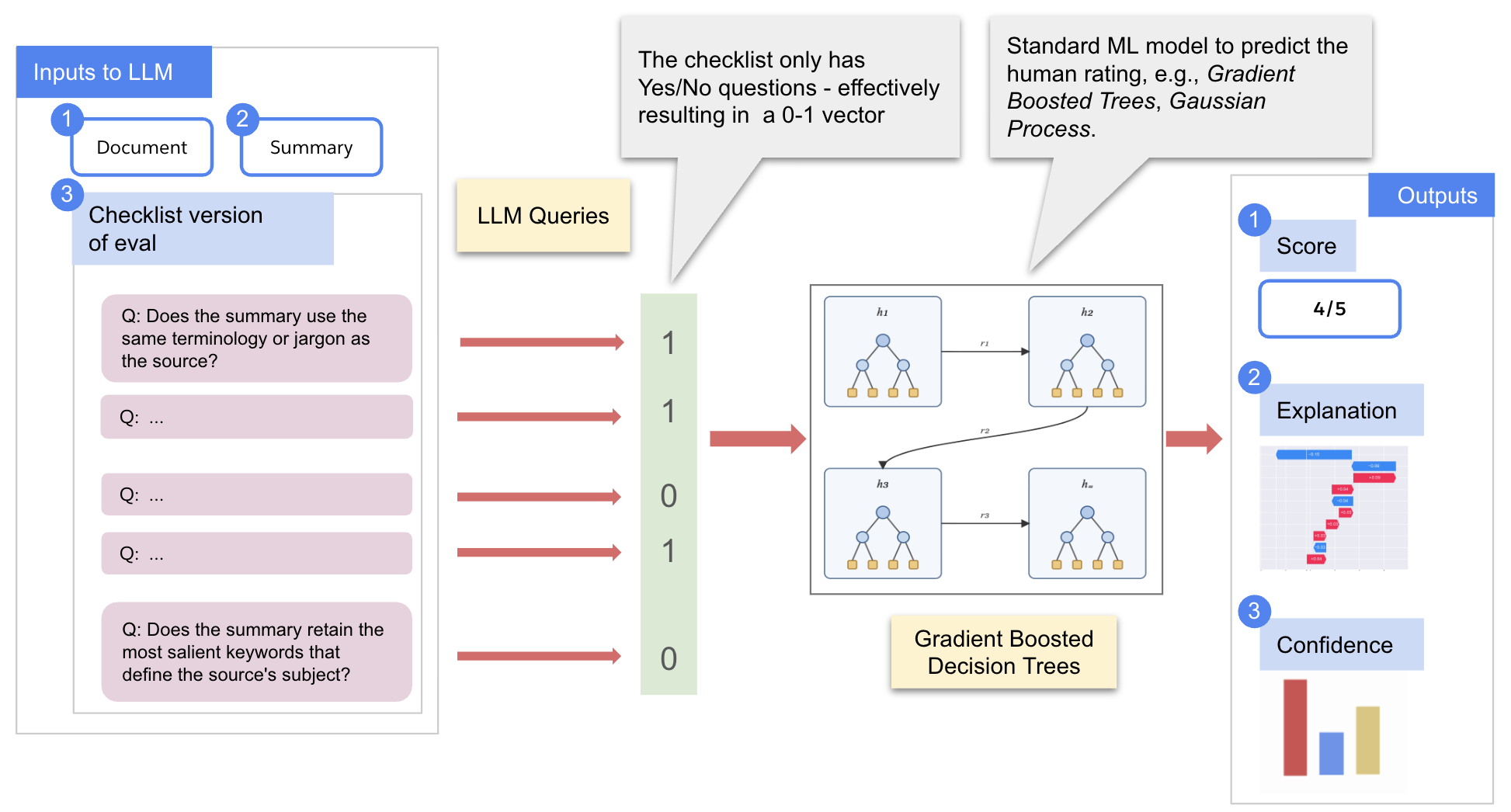}
      \caption{Our eval pipeline. A checklist of \texttt{YES}/\texttt{NO} eval questions are sent to an LLM, along with inputs for the eval. In this example, the eval assesses the quality of summarization, so a document and its corresponding summary constitute our additional inputs. The outputs from the LLM are sent to a  tabular model, e.g., \emph{Gradient Boosted Decision Trees}, to produce the final judgment score. An explanation, and some form of confidence score,  are  also produced. See \S \ref{sec:overview} for details.}
      \label{fig:pipeline}
  \end{figure}

\section{Methodology}
\label{sec:overview}

Our eval pipeline is shown in Figure \ref{fig:pipeline}. All evals are represented as a \emph{checklist} of $d$ atomic \texttt{YES}/\texttt{NO} questions. The checklist, along with eval inputs, are sent to an LLM, which produces a binary vector of responses $x \in \{0,1\}^d$. A model $f:\{0,1\}^d \mapsto \mathbb{R}$ learned earlier is used to ``aggregate'' these responses to predict a judgment $y=f(x)$. We produce two additional outputs: (1) explanations using SHAP, which help a user trace back a judgment to specific questions, and (2) a representation of confidence, either in the form of a \emph{calibrated probability} \citep{Platt1999,guo_calib,boosting_calib} or a high-confidence interval via \emph{conformal prediction} \citep{cofnromal_pred}.

The checklist and the model are created in a prior \emph{eval-authoring} step. This requires human inputs in the form of: (a) an eval prompt - we'll refer to this as the \emph{seed}  prompt, and (b) sample inputs and outputs for their eval. The step produces, as output,  (1) an eval checklist, and (2) a model $f$.

The seed prompt is decomposed into multiple simpler \texttt{YES}/\texttt{NO} questions. The idea is that (a) having targeted atomic questions  leaves little room for ambiguity, and (b) having multiple  of them acts as an ``error buffer'': even if the LLM misinterprets one question, it is unlikely that it will misinterpret every related question, which leads to a stable fraction of \texttt{YES} responses. This strategy leads to both improved \textit{agreement} and \textit{self-consistency} (empirically shown in \S \ref{sec:observations}).

We are motivated to follow the above strategy due to its success in a variety of eval settings \citep{min-etal-2023-factscore,checkeval,hdeval-liu-etal-2024-hd,li-etal-2025-dna,xu2026beyond}. We specifically use \emph{CheckEval} \citep{checkeval} because of its good results wrt inter-LLM agreement.

Since our goal also is \textit{accuracy} wrt human judgment, we use the eval input-output examples to construct our model $f$. The technique of using a model in the final step has successful precedents \citep{llmrubric,hdeval-liu-etal-2024-hd}. We use \emph{Gradient Boosted Decision Trees (GBDT)} \citep{gradboost,lightgbm} in this work, which gives us positive results (see \S \ref{sec:observations}). 

In addition to being a  powerful model family, GBDTs offer other benefits: (a) they may be easily learned for a specific  output \emph{type}, e.g., $y$ may be boolean, ordinal, categorical or continuous; (b) we use  \emph{SHAP} \citep{shap} to generate prediction explanations, which permits exact explanations for tree-based models such as GBDTs.

\section{Experiment Setup}
\label{sec:experiments}
In our experiments, we rigorously measure \emph{agreement}, \emph{self-consistency} and \emph{accuracy}, and present persuasive results for confidence scoring and explanations (these rely on standard techniques). 

\textbf{Dataset} and \textbf{ LLMs}: The \emph{SummEval} \citep{summeval} dataset (MIT License) is used for experiments. It consists of documents, their corresponding summaries (more than one summary per document), and human-provided scores per summary that assess their quality across different axes. These axes are  \emph{coherence}, \emph{consistency}, \emph{fluency} and \emph{relevance}, and we have one eval corresponding to each axis\footnote{We'll use these interchangeably  - a quality axis is assessed by an eval.}. Each summary, per  quality axis, is rated by multiple human raters, with the provided scores being   in $\{1,2,3,4,5\}$. The summaries are our measurement units, and for summary $s_i$ and an eval axis, we average the human ratings to obtain the \emph{ground truth (GT)} label $y_i \in [1,5]$.



We generate LLM scores for two additional configuration parameters :
\begin{enumerate}
    \item \textbf{LLMs}: We use $L=4$ different LLMs to measure agreement. The LLMs used are \emph{Opus 4.8} (abbreviated to \texttt{opus}), \emph{Sonnet 4.6} (\texttt{sonnet}), \emph{GPT 5.6 Sol} (\texttt{sol}), \emph{Grok 4.6} (\texttt{grok}). For \texttt{sonnet} the \emph{temperature} was set to $0$; the others don't support this parameter.
    \item Number of trials: Each eval execution is repeated $T$ times to measure statistical significance. In our experiments $T=5$.
\end{enumerate}
These additional configurations will be denoted by the indices $l$ and $t$ respectively;  the predicted scores are denoted as $\hat{y}_{ilt}$ for summary $s_i$ for a particular eval axis.


Assuming a checklist size of $d$ questions for an eval, a response vector generated for various combinations is  denoted by $x_{ilt} \in \{0,1\}^d$ .
We train one GBDT model per eval, on tuples of the form $(x_{ilt}, y_i)$ to minimize $RMSE(y_i, \hat{y}_{ilt})$ on a held-out set. Thus, for an eval, each GT label is paired with $L \times T$ inputs. Essentially, the model is trained to map from various noisy realizations of $x$ to a GT label. We use a train-conformal-test split of $50:10:40$ (the conformal split is used in \S \ref{sec:conformal_predictions}), and to prevent data leakage, we ensure that data related to  a specific $s_i$ is in exactly one split.  For additional details, please see \S \ref{sec:appendix_experiment}.

\textbf{Metrics}: As mentioned above, accuracy is measured using the RMSE score. \emph{Lower is better}.

For measuring agreement and self-consistency (note, these scores don't use GT, only $\hat{y}_{ilt}$ ), we do not use the popular \emph{Krippendorff's}-$\alpha$ score because: (a) we have multiple trials per rater/LLM, which $\alpha$ doesn't naturally accommodate, and (b)  $\alpha$ might be negative \citep{10.1162/coli.07-034-R2}, and we prefer a range of $[0,1]$ for easy interpretation.

Instead we map our  scores from a range of $[1,5]$ to $[0,1]$, and define the disagreement between two scores $a$ and $b$ as $D=|a-b|$. For an eval, this is averaged over trials for a measurement unit (here, a summary) to produce $\bar{D}_i$. For \emph{agreement}, the averaging for two given LLMs\footnote{In our setup, to claim that an output is from a specific \texttt{LLM A} means that the checklist responses $x \in \{0,1\}^d$ were obtained using \texttt{A}, and then the learned aggregator model $f$ for an eval was used to produce the final judgment.} is over their $T^2$ possible trial combinations, and for \emph{self-consistency} the $T(T-1)/2$ unique combinations for an LLM are used. Finally we report $1-\bar{D}_{i}$ as our score (\emph{agreement} or \emph{self-consistency}, depending on the averaging) for this unit. Further averaging is called out in our analyses in \S \ref{sec:observations}. 


Of course, since these scores are not chance-adjusted, they should be interpreted  jointly with \emph{accuracy}.

\textbf{Hardware}: The LLMs were accessed through an Amazon Bedrock API. All experiments were run on a M3 Max Apple Macbook Pro laptop.

\section{Results}
\label{sec:observations}
This section presents results from our setup as compared to a baseline. The latter uses seed prompts from prior literature - see \S \ref{sec:baseline_prompts}.

\begin{figure}[!ht]
      \centering
      \begin{subfigure}[t]{0.31\textwidth}
          \centering
          \includegraphics[width=\linewidth]{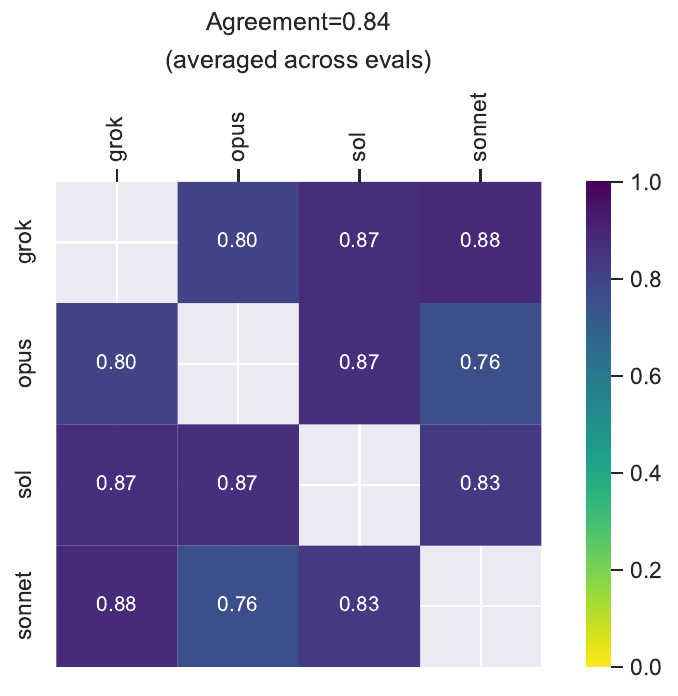}
          \caption{Baseline agreements.}
          \label{fig:left}
      \end{subfigure}
      \hfill
      \begin{subfigure}[t]{0.31\textwidth}
          \centering
          \includegraphics[width=\linewidth]{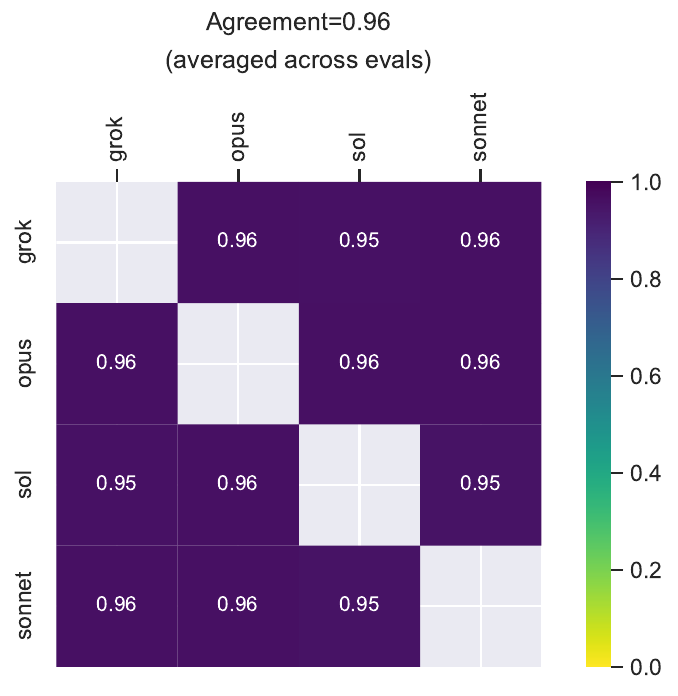}
          \caption{Agreements - all LLMs in train/test.}
          \label{fig:right}
      \end{subfigure}
            \hfill
      \begin{subfigure}[t]{0.31\textwidth}
          \centering
          \includegraphics[width=\linewidth]{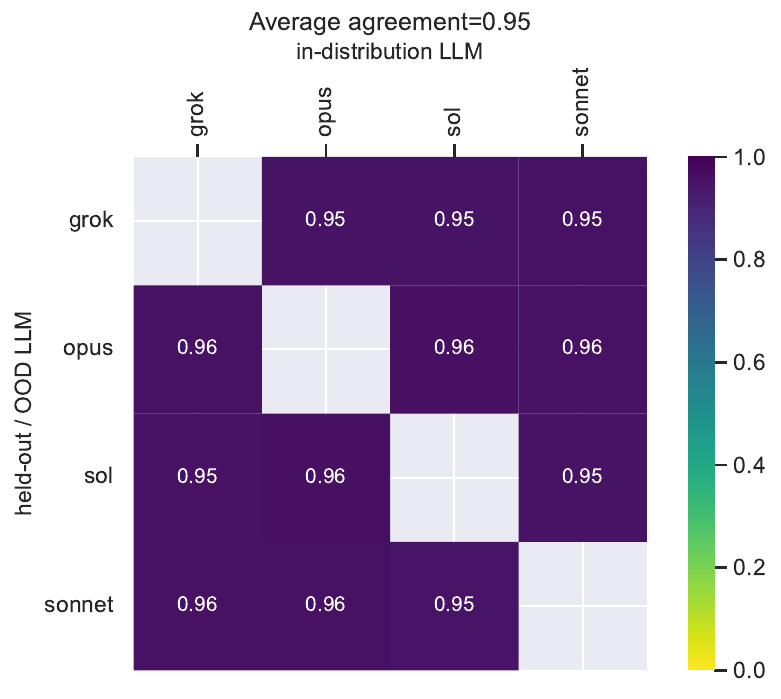}
          \caption{Agreements - with one held-out LLM.}
          \label{fig:right}
      \end{subfigure}
      \caption{LLM agreements on held-out test data (higher is better). (a) shows baseline agreements, (b) and (c) are from our setup. In (b) data from all LLMs were present both in  train and test splits. In (c), for a row, data from the LLM indexing the row was absent in train, but agreements are reported with the other LLMs on a held-out set. See text for details.}
      \label{fig:agreements}
  \end{figure}

\textbf{Agreement}: The average agreement for a pair of LLMs is shown as $4 \times 4$ heatmap in Figure \ref{fig:agreements}. A particular cell further averages $1-\bar{D}_{i}$  over all summaries and evals in a common held-out set. Figure \ref{fig:agreements}(a) shows baseline results, and both (b) and (c) come from our setup. 

In Figure \ref{fig:agreements}(b), the aggregator model $f$ has seen data from all LLMs during training, and a cell indexed by $[A,B]$ compares the agreement between LLMs $A$ and $B$ on a held-out set. We notice that (b) produces much higher scores than (a), and the overall average (in the plot title) improves: $0.84 \rightarrow 0.96$. 

Perhaps more interesting is Figure \ref{fig:agreements}(c). Here \emph{an LLM is held-out} as well, and we cycle through all the LLMs, denoted by rows. For example, the first row indicates that the model $f$ only saw data from \texttt{opus}, \texttt{sol} and \texttt{sonnet} in the training data, but the agreement is computed between each of these and \texttt{grok}, on a held-out set. Even here, we note high scores for all pairs.


\textbf{Self-consistency}: The self-consistency scores on the baseline prompts, and then on the output from our framework, are shown in Table  \ref{tab:consistency-per-llm}. The score for an LLM are averaged over summaries and evals. Here, the scores look good to start with, although  \texttt{Sol} and \texttt{Grok} seem to benefit most from our approach, and the overall average  slightly improves.


 \begin{table}[h]
    \centering
    \begin{tabular}{lccccc}
    \toprule
     & \texttt{grok} & \texttt{opus} & \texttt{sol} & \texttt{sonnet} &
  {Average} \\
    \midrule
    Baseline   & 0.94 & 0.97 & 0.96 & 0.99 & 0.97 \\
    Our setup  & 0.97 & 0.99 & 0.98 & 0.99 & 0.98 \\
    \bottomrule
    \end{tabular}
    \vspace{0.5em}
    \caption{Self-consistency averaged across all evals, per LLM. Higher is better.}
    \label{tab:consistency-per-llm}
    \end{table}
    
\begin{figure}[!ht]
      \centering
      \includegraphics[width=1\linewidth]{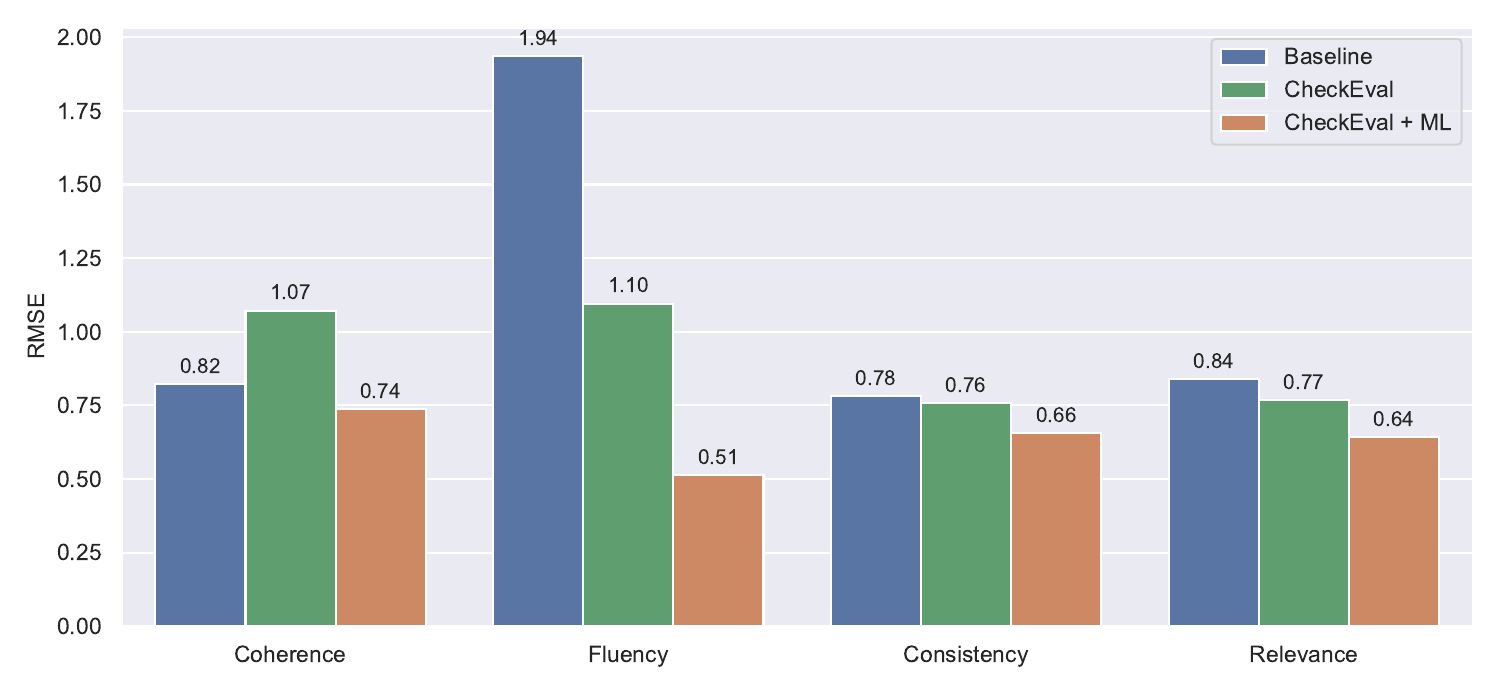}
      \caption{Eval accuracy measured using RMSE (lower is better).}
      \label{fig:accuracy_rmse}
  \end{figure}

  \textbf{Accuracy}: Accuracy scores, per eval, are shown as grouped bar plots in Figure \ref{fig:accuracy_rmse}. Each group shows a bar for the baseline RMSE (labeled \texttt{Baseline}), aggregation via just reporting the fraction of \texttt{YES} answers (labeled \texttt{CheckEval}), and the output from the entire pipeline  (labeled \texttt{CheckEval+ML}). 
  
  We show the \texttt{CheckEval} bars to emphasize that reducing variability - which CheckEval is effective at when the fraction of \texttt{YES}es are reported as the final score (as reported in the original work \citet{checkeval}) is \emph{insufficient to obtain good accuracy}. In fact, for the \emph{coherence} eval, \texttt{CheckEval} is worse than \texttt{Baseline}. In all cases, \texttt{CheckEval+ML} produces the lowest RMSE score.

\section{Explanations}
\label{sec:explanations}
It is often not enough to know why a test produced a certain result; the ability to trace an outcome back to causes provides an actionable path to improving a product feature. For a typical LLM eval driven by one prompt, since both the prompt and input might be long, pinpointing an exact fragment in the input, and the corresponding violated rubric in the eval prompt, is much harder. The simple strategy of \emph{asking} the LLM to explain its judgment has been shown to be  unreliable:  LLMs tend to favor \emph{plausibility} over \emph{faithfulness} \citep{NEURIPS2023_ed3fea90,madsen-etal-2024-self,chen2025reasoningmodelsdontsay}.

\begin{figure}[!ht]
      \centering
      \includegraphics[width=1\linewidth]{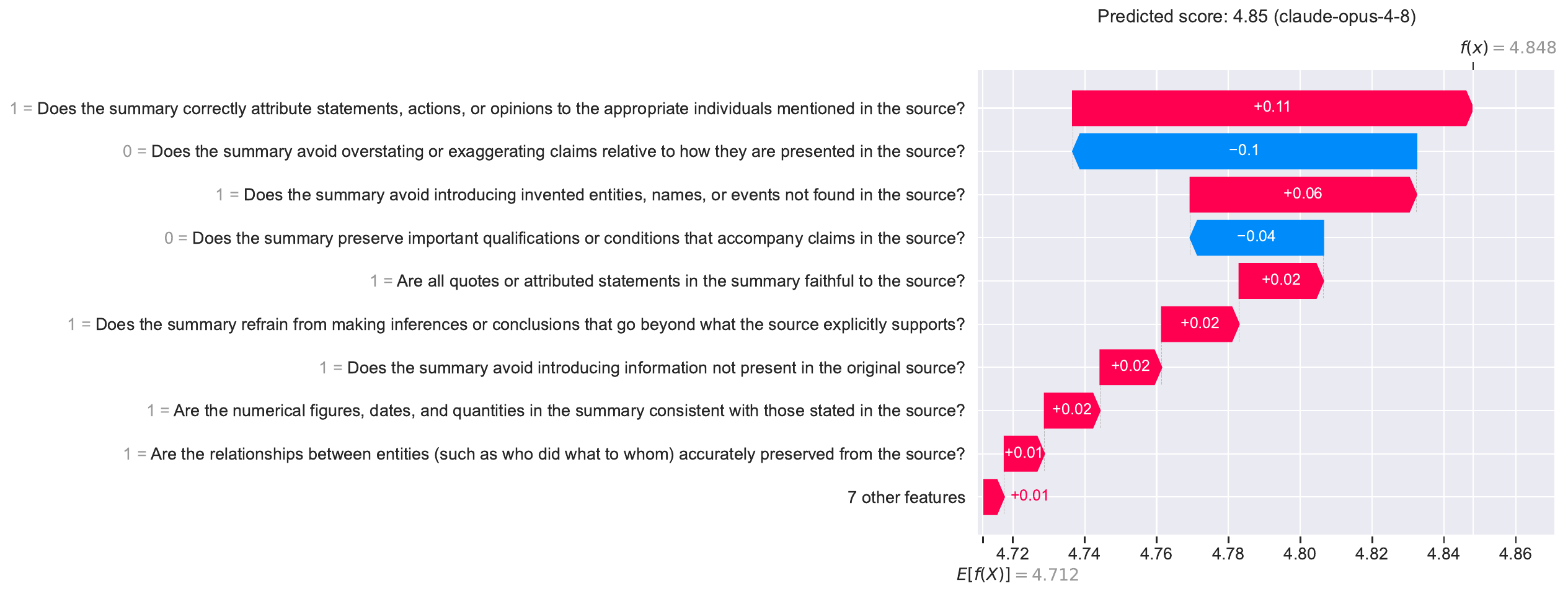}
      \caption{A standard SHAP waterfall chart for a particular input - for the \emph{consistency} eval, on \texttt{opus} - showing influences of various questions in a checklist, in decreasing order from top-to-bottom.}
      \label{fig:shap_local}
  \end{figure}

Our framework is able to sidestep this issue:
\begin{itemize}
    \item Because our eval is distilled into a checklist, there are specific contributing questions to point back to.
    \item Using a tabular model as the aggregator allows us to use a popular technique like SHAP to identify questions most influential to a specific outcome. Additionally for GBDTs,  \emph{TreeSHAP} \citep{treeshap} provides exact SHAP attribution values. Figure \ref{fig:shap_local} shows a \emph{waterfall} plot\footnote{ \url{https://shap.readthedocs.io/en/latest/example_notebooks/api_examples/plots/waterfall.html}} of question influences, produced by SHAP.
\end{itemize}

\section{Prediction Confidence}
\label{sec:conformal_predictions}

We discuss how we quantify uncertainty, which arises due to various factors such as variability across LLMs, non-determinism, and eval inputs that are poorly represented in the training data.

While this is convenient with a model like \emph{Gaussian Process} \citep{gaussian_process_book}, as mentioned earlier, we prefer a tree-based model due to high-fidelity SHAP explanations. We use conformal predictions (CP) here to report prediction intervals that are likely to contain the true outcome $90\%$ of the time. Being model-agnostic, CP also makes this step invariant to future model changes. We use the \emph{residual-normalized} version \citep{Lei03072018_residual,mapie} to make the intervals input-adaptive. 

\begin{figure}[!ht]
      \centering
      \begin{subfigure}[t]{0.45\textwidth}
          \centering
          \includegraphics[width=\linewidth]{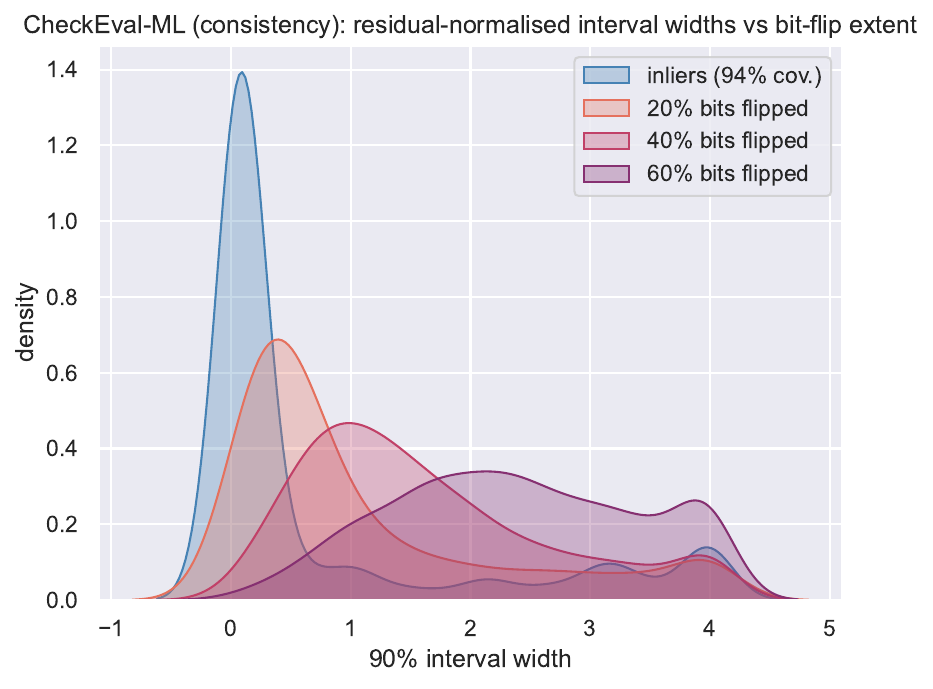}
          \caption{\emph{consistency} eval}
          \label{fig:baseline_consistency}
      \end{subfigure}
      \hspace{0.02\textwidth}
      \begin{subfigure}[t]{0.45\textwidth}
          \centering
          \includegraphics[width=\linewidth]{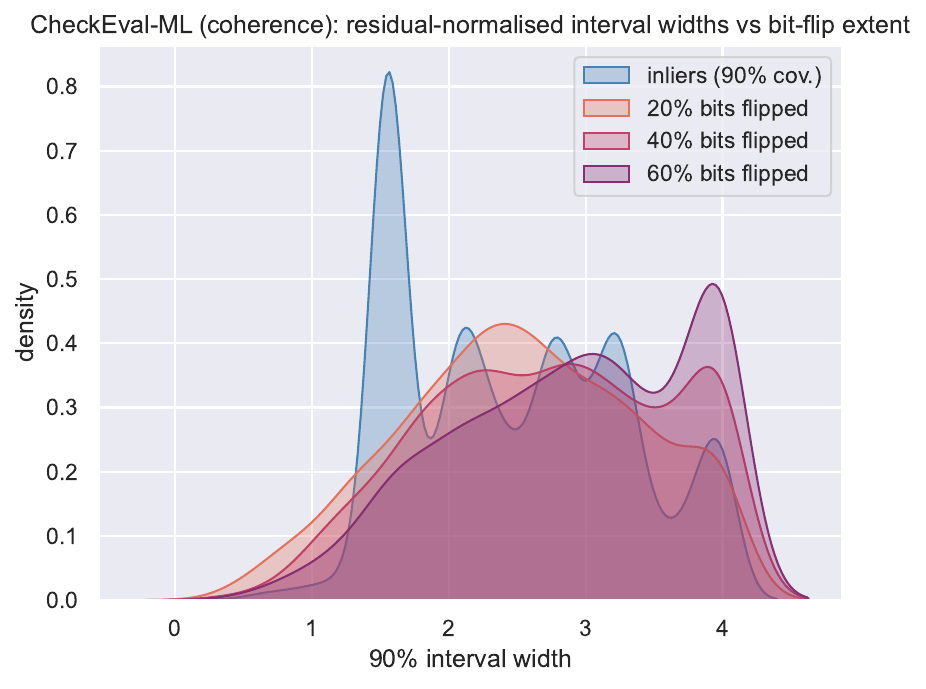}
          \caption{\emph{coherence} eval}
          \label{fig:right_consistency}
      \end{subfigure}
      \caption{Distribution of interval widths on actual (inliers) vs synthetic noisy data. }
      \label{fig:confidence}
  \end{figure}

To validate, we generate synthetic data   from valid $x_{ilt} \in \{0,1\}^d$ vectors by flipping $b \in \{0.2, 0.4, 0.6\}$ fraction of bits, and visualize  the distribution of the interval widths; shown in Figure \ref{fig:confidence}, for  the (a) \emph{consistency}, and (b) \emph{coherence} evals. The ``inliers'' KDE plot is on the test data, and is expected to have relatively small widths; the legend shows its empirical coverage. In both cases, we observe that higher values $b$ lead to wider intervals (expected behavior). See \S \ref{sec:appendix_cp} for more plots.


\section{Limitations and Conclusion}
\label{sec:conclusions}
In this work we presented a reliable eval pipeline that meets multiple practical requirements,  listed in \S \ref{sec:practical_eval_pipeline}. We think this work is important in bringing academic research into the industry; even if the exact pipeline isn't used, the general architecture and specific components may be built upon. An obvious limitation of this work is that it is validated on a single dataset for the task of summarization, against proprietary LLMs; something that the authors are working  to address. 

\section{Acknowledgements}
This work was possible with the support of the \emph{Q3 Salescloud} team in Salesforce, specifically Ronnie Fong, Nitin Jain and William Hackett. We are grateful to Harsha Medikonda, Prasad Maderamitla, Hinita Patel, Aaron Quan and Soumya Mittapalli, for  pointing us to multiple real-world challenges. We also would like to thank Pratik Gupte, Kimberley Lee and Daniel May for various technical suggestions.

\newpage
\bibliographystyle{plainnat}
\bibliography{refs}


\appendix
\section{Additional Experiment Details}
\label{sec:appendix_experiment}

\subsection{Baseline Prompts}
\label{sec:baseline_prompts}

We use the G-Eval \citep{geval} prompts without the autogenerated chain-of-thoughts evaluation steps as the baseline prompts in our experiment.
\newtcolorbox{promptbox}[1]{breakable,enhanced,colback=gray!5,colframe=black!60,
    boxrule=0.5pt,arc=2pt,fonttitle=\bfseries\small,title={#1}}

  \begin{promptbox}{Coherence}
  \begin{Verbatim}[fontsize=\small,breaklines=true,breakanywhere=true]
  You will be given one summary written for a news article.

  Your task is to rate the summary on one metric.
  
  Please make sure you read and understand these instructions carefully.
  Please keep this document open while reviewing, and refer to it as needed.

  Evaluation Criteria:

  Coherence (1-5) - the collective quality of all sentences. We align this
  dimension with the DUC quality question of structure and coherence whereby
  "the summary should be well-structured and well-organized. The summary 
  should not just be a heap of related information, but should build from 
  sentence to a coherent body of information about a topic."

  Source Text:

  {{Document}}

  Summary:

  {{Summary}}

  Evaluation Form (scores ONLY):

  - Coherence:
  \end{Verbatim}
  \end{promptbox}
  \begin{promptbox}{Consistency}
  \begin{Verbatim}[fontsize=\small,breaklines=true,breakanywhere=true]
  You will be given a news article. You will then be given one summary 
  written for this article.

  Your task is to rate the summary on one metric.

  Please make sure you read and understand these instructions carefully.
  Please keep this document open while reviewing, and refer to it as needed.

  Evaluation Criteria:

  Consistency (1-5) - the factual alignment between the summary and the
  summarized source. A factually consistent summary contains only statements
  that are entailed by the source document. Annotators were also asked to
  penalize summaries that contained hallucinated facts.

  Source Text:

  {{Document}}

  Summary:

  {{Summary}}

  Evaluation Form (scores ONLY):

  - Consistency:
  \end{Verbatim}
  \end{promptbox}

  \begin{promptbox}{Fluency}
  \begin{Verbatim}[fontsize=\small,breaklines=true,breakanywhere=true]
  You will be given one summary written for a news article.

  Your task is to rate the summary on one metric.

  Please make sure you read and understand these instructions carefully.
  Please keep this document open while reviewing, and refer to it as needed.

  Evaluation Criteria:

  Fluency (1-5): the quality of individual sentences. Drawing again from 
  the DUC quality guidelines, sentences in the summary "should have no 
  formatting problems, capitalization errors or obviously ungrammatical 
  sentences (e.g., fragments, missing components) that make the text 
  difficult to read."

  Example:
  
  Source Text:

  {{Document}}

  Summary:

  {{Summary}}

  Evaluation Form (scores ONLY):

  - Fluency:
  \end{Verbatim}
  \end{promptbox}

  \begin{promptbox}{Relevance}
  \begin{Verbatim}[fontsize=\small,breaklines=true,breakanywhere=true]
  You will be given one summary written for a news article.

  Your task is to rate the summary on one metric.

  Please make sure you read and understand these instructions carefully.
  Please keep this document open while reviewing, and refer to it as needed.
  
  Evaluation Criteria:

  Relevance (1-5) - selection of important content from the source. The 
  summary should include only important information from the source document. 
  Annotators were instructed to penalize summaries which contained 
  redundancies and excess information.

  Example:

  Source Text:

  {{Document}}

  Summary:

  {{Summary}}

  Evaluation Form (scores ONLY):

  - Relevance:
  \end{Verbatim}
  \end{promptbox}

\subsection{Generated Checklists}

  \begin{table}[h]
  \centering
  \begin{tabular}{lccc}
  \hline
  \textbf{Dimension} & \textbf{\# Sub-dimensions} & \textbf{\# Seed Questions} & \textbf{\# Final Questions} \\
  \hline
  Coherence   & 3 & 3 & 20 \\
  Consistency & 3 & 3 & 16 \\
  Fluency     & 4 & 4 & 24 \\
  Relevance   & 5 & 5 & 21 \\
  \hline
 
  \end{tabular}
  \vspace{10pt}
  \caption{Checklist size per SummEval dimension. The sub-dimensions and seed questions were taken from CheckEval \citep{checkeval}. We used \emph{Opus 4.8} to generate the final questions.}
  \label{tab:summeval-checklists}
  \end{table}

\subsection{GBDT Hyperparameters}

The GBDT in \S \ref{sec:experiments} was learned using cross-validation of hyperparameters from the following search space:
 \begin{verbatim}
    "n_estimators": [20, 50, 100, 200]
    "learning_rate": [0.001, 0.01, 0.1, 1.0]
    "max_depth": [3, 5, 10]
    "min_child_samples": [5, 10, 20, 40]
    "reg_lambda": [0.0, 1.0, 5.0]
  \end{verbatim}

\subsection{Additional Conformal Prediction Results}
\label{sec:appendix_cp}
\begin{figure}[!ht]
      \centering
      \begin{subfigure}[t]{0.48\textwidth}
          \centering
          \includegraphics[width=\linewidth]{assets/confidences_conformal_prediction/interval_widths_checkevalML_consistency.pdf}
          \caption{\emph{consistency} eval}
          \label{fig:baseline_consistency}
      \end{subfigure}
      \hspace{0.02\textwidth}%
      \begin{subfigure}[t]{0.48\textwidth}
          \centering
          \includegraphics[width=\linewidth]{assets/confidences_conformal_prediction/interval_widths_checkevalML_coherence.pdf}
          \caption{\emph{coherence} eval}
          \label{fig:right_consistency}
      \end{subfigure}

      \vspace{1em}   

      \begin{subfigure}[t]{0.48\textwidth}
          \centering
          \includegraphics[width=\linewidth]{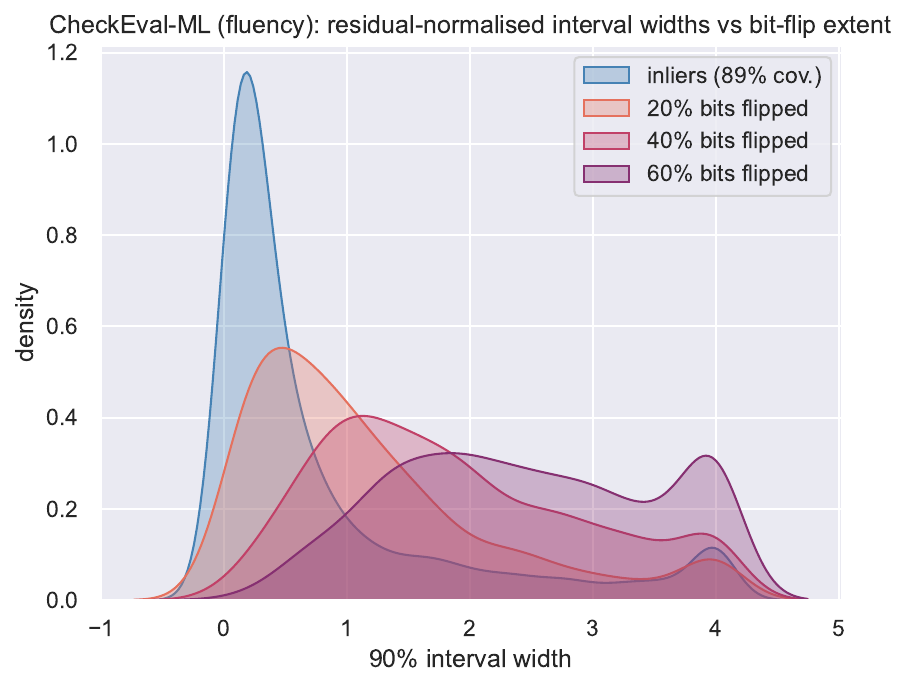}
          \caption{\emph{fluency} eval}
          \label{fig:dummy_bl}
      \end{subfigure}
      \hspace{0.02\textwidth}%
      \begin{subfigure}[t]{0.48\textwidth}
          \centering
          \includegraphics[width=\linewidth]{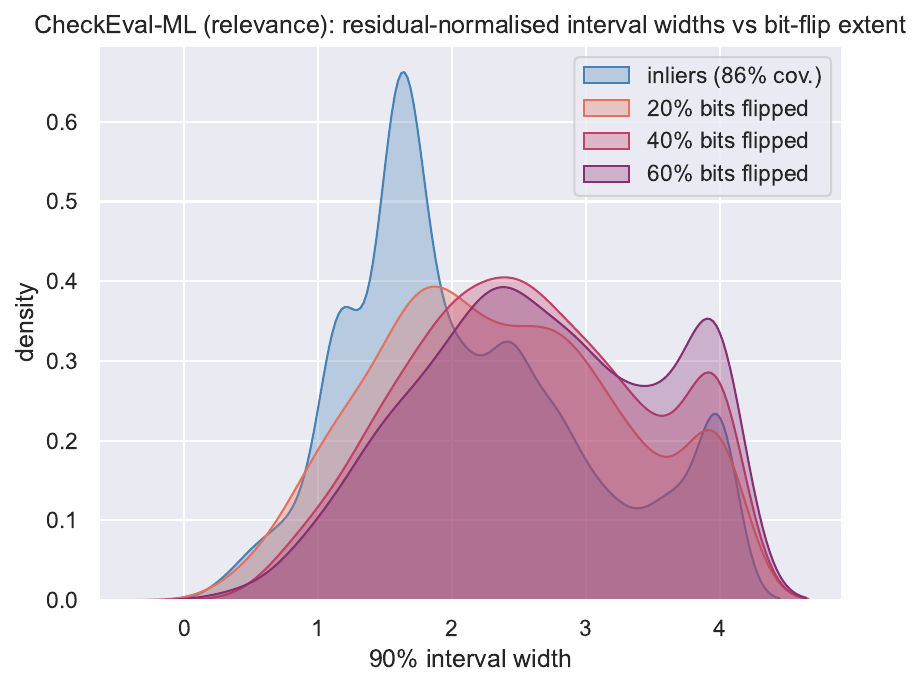}
          \caption{\emph{relevance} eval}
          \label{fig:dummy_br}
      \end{subfigure}

      \caption{Distribution of interval widths for points in the data or ``inliers'' and synthetic noisy data, which is constructed by flipping bit for different fractions of bits $b \in \{0.2, 0.4, 0.6\}$. The empirical coverage of the inliers is shown in the legend, which may be observed to be quite close to intended coverage of $90\%$.}
      \label{fig:apendix_complete_cp}
  \end{figure}

Continuing from \S \ref{sec:conformal_predictions}, we present the plots for all evals in Figure \ref{fig:apendix_complete_cp}.

\end{document}